\documentclass[letterpaper, 10 pt, conference]{ieeeconf}  

\IEEEoverridecommandlockouts                              

\usepackage{amsmath} 

\usepackage{balance} 
\usepackage{algorithm}
\usepackage{algpseudocode}
\usepackage{tikz}
\usepackage{booktabs}
\usepackage{graphicx}
\usepackage{caption}
\usepackage{subcaption}
\usepackage{lineno}
\usepackage[noadjust]{cite}
\def\checkmark{\tikz\fill[scale=0.4](0,.35) -- (.25,0) -- (1,.7) -- (.25,.15) -- cycle;}

\title{\LARGE \bf
Heterogeneous Coalition Formation and Scheduling \\
with Multi-Skilled Robots}

\author{Ashay Aswale and Carlo Pinciroli
\thanks{All the authors are with the Dept. of Robotics Engineering, Worcester Polytechnic Institute, Worcester, MA, USA (email: {\tt\small \{asaswale, cpinciroli\}@wpi.edu}).}}

\begin{document}

\maketitle
\thispagestyle{empty}
\pagestyle{empty}

\begin{abstract}

We present an approach to task scheduling in heterogeneous multi-robot systems. In our setting, the tasks to complete require diverse skills. We assume that each robot is multi-skilled, i.e., each robot offers a subset of the possible skills. This makes the formation of heterogeneous teams (\emph{coalitions}) a requirement for task completion. We present two centralized algorithms to schedule robots across tasks and to form suitable coalitions, assuming stochastic travel times across tasks. The coalitions are dynamic, in that the robots form and disband coalitions as the schedule is executed. The first algorithm we propose guarantees optimality, but its run-time is acceptable only for small problem instances. The second algorithm we propose can tackle large problems with short run-times, and is based on a heuristic approach that typically reaches 1x-2x of the optimal solution cost. 

\end{abstract}

\section{Introduction}
\label{sec:1}

The parallelism offered by multi-robot systems is a natural fit for missions in which large numbers of tasks must be achieved as quickly as possible \cite{brambilla2013swarm}. In realistic settings, each task requires robots with specific \emph{skills}, such as specific sensors, actuators, or computational resources. However, as the diversity of the tasks increases and the set of required skills grows, it becomes infeasible to envision swarms of identical robots that can interchangeably perform any task. Rather, specialization and redundancy become desirable due to better scale economy and expected long-term resilience \cite{ramachandran2019resilience,prorok2021beyond}.

The goal of this paper is to contribute to realizing this vision. In our setting, a heterogeneous swarm of multi-skilled robots must perform a set of tasks as quickly as possible. We assume that, due to the diversity of the tasks, the robot must form \emph{coalitions}, i.e., teams of robots that, combined, offer the required skills for each task to be completed \cite{barton2008coalitionSocial,capezzuto2020coalitions,capezzuto2021schedulingCoalition}. In addition to heterogeneity, our problem setting has two crucial features: \emph{(i)} the diversity of the tasks also requires the coalitions to be \emph{dynamically} formed and disbanded on a per-task basis; and \emph{(ii)} all the required robots in a coalition must be present at the same time for the task to progress. These two features imply that, along with the combinatorial problem of forming coalitions, the robots must also \emph{schedule} the optimal task agenda in a coherent manner.

The key difference between our work and existing works is that we consider \emph{simultaneously} multi-skill coalition formation and multi-robot task scheduling for a complete coverage problem.
With reference to the Korsah \textit{et al.} taxonomy \cite{korsah2013taxonomy}, this problem is an instance of cross-schedule dependencies (XD), single-task robots (ST), multi-robot tasks (MR), and time-extended assignment (TA). In addition, to make our problem setting more realistic, we enrich our formulation with \emph{stochastic travel times} across tasks \cite{nam2016optimalAssignment}.

We study two centralized algorithms to solve this problem. The first algorithm is optimal, but it scales poorly with the number of tasks, robots, and skills. In contrast, the second method scales to hundreds of tasks, robots, and skills. Even though the latter method offers no optimality guarantees, we empirically show that its performance is within a factor of 2 with respect to the optimum.


\begin{figure}[t!]
    \centering
    \includegraphics[width=0.45\textwidth]{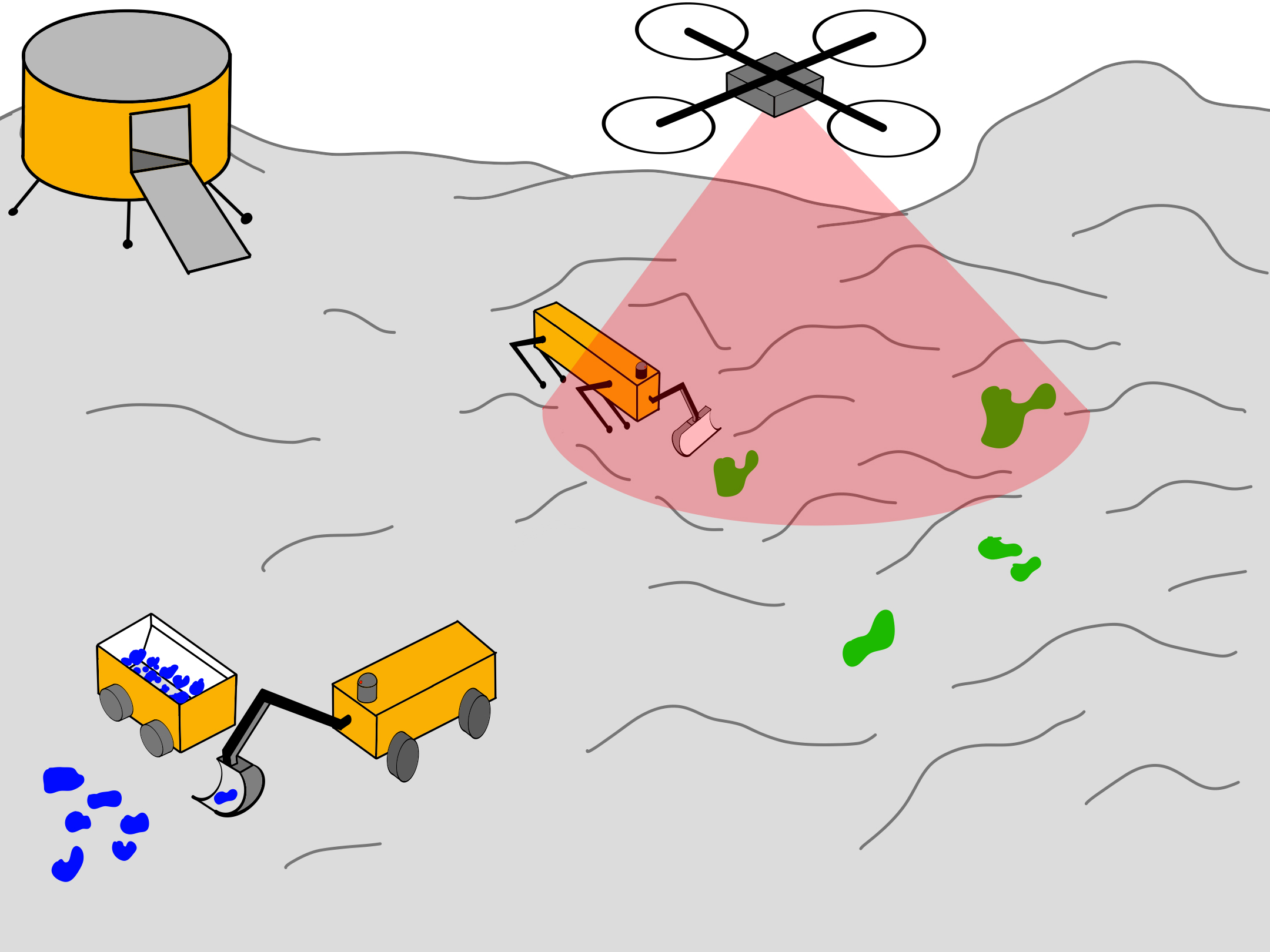}
    \caption{A heterogeneous swarm of multi-skilled robots forming coalitions. \textmd{The tasks in this diagram are excavation of blue resources (bottom left), and green material sampling (center). The excavation task requires a resource sensor, an excavating arm, and a bucket to carry resources. Material sampling requires a scanner to find and locate the sample, and an arm on a legged robot.}}
    \label{fig:banner_image}
\end{figure}

The rest of this paper is structured as follows. In Sec. 2 we survey related work on coalition formation and multi robot scheduling. In Sec. 3 we discuss the problem formulation for both of our approaches. In Sec. 4 we analyze the results of both the methods and compare them with each other. We conclude the paper in Sec. 5.

\section{Related work}
\label{sec:2}



The problems of task scheduling and coalition formation have received wide attention in the literature. 
We identified several axes to categorize relevant work in Table \ref{tab:backg}.


Several works study coalition formation without scheduling. Rahwan \textit{et al.} \cite{rahwan2007coalitionAlgos} and Guo \textit{et al.} \cite{guo2020repairs} focus on \textit{homogeneous} coalition formation. Rahwan  \textit{et al.} \cite{rahwan2007coalitionAlgos} considers \textit{static} coalitions in which coalitions, once formed, are kept constant throughout the experiment. Guo \textit{et al.} \cite{guo2020repairs} compare \textit{static} and \textit{dynamic} coalition formation, in which the robots are allowed to change coalition from task to task.  

Recently, several researchers considered coalition formation of heterogeneously skilled robots  \cite{hu2022coverage,huang2020inner,lin2021deployment}. However, they focus on coverage and connectivity problems, neglecting scheduling aspects. Similarly, Barton \textit{et al.} \cite{barton2008coalitionSocial} study coalition formation in a network of \textit{heterogeneously skilled} agents. 
At the opposite end of the spectrum, Parker \textit{et al.} \cite{parker2016exploiting} and Visser \textit{et al.} \cite{visser2018integrating}  study search-and-rescue scenarios with a heterogeneous set of robots and focus on scheduling them without coalition formation.

Significant effort has been devoted to combining coalition formation with task scheduling \cite{guerrero2017deadlines,ramchurn2010coalitionSpacialTemporal,koes2006cocoa,capezzuto2020coalitions,capezzuto2021schedulingCoalition}.
However, all of these works assume the robots to be homogeneously skilled.





Some of the closest research to this paper studies coalition formation with heterogeneous robots that also produces an optimal plan with cross-scheduling dependencies \cite{Leahy2022scalable,lippi2021HRI,korsah2012xbots,mansfield2021mrscheduling}. 
However, these works do not consider \textit{multi-skilled} robots; rather, each robot is a specialist of a specific skill.

Prorok \textit{et al.} \cite{prorok2017impact,prorok2016fastRedistribution} and Kosak \textit{et al.} \cite{kosak2018multipotent} address multi-skilled robots. In the work of Prorok \textit{et al.}, \cite{prorok2017impact,prorok2016fastRedistribution}, the robots offer a subset of the possible skills, whereas Kosak \textit{et al.} \cite{kosak2018multipotent} allow ``multipotent'' robots to modify themselves and adapt to the task at hand. However, all these works focus on coalition formation without a scheduling component. 

Amador \textit{et. al.} \cite{amador2014dynamic, tkach2021towards} addressed a comparable issue, the `Law enforcement problem' (\textit{LEP}), which assigns police officers to tasks with unknown locations, arrival times, and importance levels. Although \textit{LEP} considers cross-scheduling dependencies for multi-skilled agents, their problem statement differs significantly from ours. Their approach concerns time-sensitive tasks with various importance levels and agents who can abandon or ignore tasks if they do not offer a higher utility. In contrast, our problem assumes all tasks are equally important and not time-sensitive. Moreover, our agents cannot abandon any tasks, and all tasks must be completed.

 We are also interested in the presence of stochastic aspects in the problem statement. The works considered so far lack such a component, but recent research has started to include it. Nam \textit{et al.}  \cite{nam2016optimalAssignment}, for example, include stochastic travel times in multi-agent task scheduling; however, their work does not require coalitions formation. In the literature on coalitions, stochastic aspects concern resilience and reconfiguration \cite{mayya2021resilient,ramachandran2019resilience,ramachandran2021resilient}, without a scheduling component.







\begin{table}[t]
\caption{Comparison of the related work to the present work. Legend: Coalt: Coalition, Heter: Heterogeneous set of robots, Sched: Scheduling, Stoch: Stochasticity of any nature, M-Skill: Multi-skilled robots.}
\begin{tabular}{cccccc}
 \toprule
 \textbf{Work} & \textbf{Coalt} & \textbf{Heter} & \textbf{Sched} & \textbf{Stoch} & \textbf{M-Skill} \\ 
\midrule
\cite{rahwan2007coalitionAlgos,guo2020repairs} & \checkmark & & & & \\
\cite{barton2008coalitionSocial,hu2022coverage,huang2020inner,lin2021deployment} & \checkmark & \checkmark &&&\\
\cite{guerrero2017deadlines,ramchurn2010coalitionSpacialTemporal,koes2006cocoa,capezzuto2020coalitions,capezzuto2021schedulingCoalition} & \checkmark & & \checkmark & & \\
\cite{parker2016exploiting,visser2018integrating} & & \checkmark & \checkmark & & \\
\cite{Leahy2022scalable,lippi2021HRI,korsah2012xbots,mansfield2021mrscheduling} & \checkmark & \checkmark & \checkmark & & \\
\cite{mayya2021resilient,ramachandran2019resilience,ramachandran2021resilient} & \checkmark & \checkmark & & \checkmark & \\
\cite{nam2016optimalAssignment} & & & \checkmark & \checkmark & \\
\cite{prorok2017impact,prorok2016fastRedistribution,kosak2018multipotent} & \checkmark & \checkmark & & & \checkmark\\
\cite{amador2014dynamic, tkach2021towards} & \checkmark & \checkmark & \checkmark & & \checkmark\\
\textit{this work} & \checkmark & \checkmark & \checkmark & \checkmark & \checkmark\\
 \bottomrule
\end{tabular}
\label{tab:backg}
\end{table}


%
\section{Approach}
\label{sec:3}

\subsection{Preliminaries}



We consider an environment in which a set of $m$ tasks is scattered. Each task requires a specific set of skills. A task can require any number of skills. In total, across all tasks, $l$ skills are required; some might be in higher demand than others. The mapping between tasks and required skills is captured by the binary matrix $R$, whose element $r_{js}$ is 1 if task $j$ requires skill $s$ and 0 otherwise.

A swarm of $n$ robots is deployed in the environment and must perform the tasks as quickly as possible. The robots may start from the same `depot' or be scattered throughout the environment. We assume that each robot offers a subset of the possible skills, and the robots combined offer all the required skills for the tasks to be completed. We restrict each robot to have a maximum of $l/2$ skills. The binary matrix $Q$ encodes the mapping between robots and skills. An element $q_{is}$ is 1 if robot $i$ possesses skill $s$ and 0 otherwise. 

\subsection{Optimal Formulation}
We first discuss a Linear Program that produces the optimal solution to the problem. We use a binary assignment tensor  ($X$) for the formulation. In this tensor, if an element $x_{ijk}$ is 1, then robot $i$ attends task $k$ right after task $j$; otherwise the element is 0. We refer to the robot’s location at the start and the end of the experiment as ``task 0'' and ``task $m+1$'' respectively. 

We denote with $T$ the cost tensor that stores the travel time from one task to another. Because the robots might start from different locations, the travel cost from task 0 to the other tasks differs from robot to robot. For this reason, $t_{ijk}$ denotes the travel time for robot $i$ to navigate from task $j$ to task $k$. Travel times at this stage are deterministic; we will explain how to handle stochastic times in Sec. \ref{subsec:stoch_time}.

\begin{table}[!t]
\caption{Table of parameters and variables}
\begin{tabular}{ p{0.06\textwidth} p{0.375\textwidth} } 
 \toprule
 \textbf{Symbol} & \textbf{Description}\\ 
\midrule
$l$ & Total number of skills in a configuration\\
$m$ & Total number of tasks in a configuration\\
$n$ & Total number of robots in a configuration\\
$Q$ & Binary matrix for mapping between robots and skills\\
$R$ & Binary matrix for mapping between tasks and requirements\\
$T$ & Cost tensor\\
$T^s$ & Matrix for the stochastic buffer time for travel\\
$T^e$ & Vector for execution time of the tasks\\
$X$ & Binary assignment tensor\\
$Y$ & Matrix for arrival times of the robots at tasks\\
$Y^{\max}$ & Vector for the tasks' execution start times\\
$Z$ & Matrix for how many robots offer each skill\\
$Z^b$ & Matrix to denote excess skills at the tasks\\
\bottomrule
\end{tabular}
\label{tab:params_and_vars}
\end{table}


\subsubsection{Valid schedule generation}

We now discuss the constraints necessary to generate valid schedules for the robots.
Every robot $i$ must start from task 0 (initial location) exactly once (Eq. \eqref{eq:start-any}) and finish its schedule at task $(m+1)$ (final location) exactly once (Eq. \eqref{eq:any-end}). Therefore, task 0 is exit-only (Eq. \eqref{eq:any-start}) and task $(m+1)$ is entry-only (Eq. \eqref{eq:end-any}).

\begin{align}
\forall i &\quad \sum^{m+1}_{k=1}x_{i0k} = 1 \label{eq:start-any}\\
\forall i &\quad \sum^{m}_{j=0}x_{ij(m+1)} = 1 \label{eq:any-end}\\
\forall i &\quad \sum^{m+1}_{j=1}x_{ij0} = 0 \label{eq:any-start}\\
\forall i &\quad \sum^{m}_{k=0}x_{i(m+1)k} = 0 \label{eq:end-any}
\end{align}

For what concerns the other tasks $[1,m]$, we impose that a task cannot appear twice in a robot's schedule. To this effect, a robot can enter a task $k$ at most once (Eq. \eqref{eq:max_arrive}) and exit it at most once (Eq. \eqref{eq:max_leave}). In addition, if a robot has visited a task $j$, it must leave it, and it cannot leave it without first visiting it (Eq. \eqref{eq:in_out}). Finally, robots cannot dwell at a task after visiting it (Eq. \eqref{eq:not_stay}).
\begin{align}
\forall i \forall k \backslash \{0,m+1\} &\quad  \sum^{m}_{j=0}x_{ijk} \leq 1 \label{eq:max_arrive} \\
\forall i \forall j \backslash \{0,m+1\} &\quad  \sum^{m+1}_{k=1}x_{ijk} \leq 1 \label{eq:max_leave} \\
\forall i \forall j \backslash \{0,m+1\} &\quad  (\sum^{m}_{k=0}x_{ikj} = \sum^{m+1}_{k=1}x_{ijk}) \label{eq:in_out} \\
\forall i \forall j  &\quad  x_{ijj} = 0 \label{eq:not_stay}
\end{align}


These constraints could result in schedules where a robot travels along multiple different paths at the same time. 
There might be one valid path, starting from task $0$ and ending at task $m+1$, and an invalid path, looping between three or more tasks. To solve this, we add ``lazy constraints'' that reject a candidate solution if it contains such loops. The algorithm to detect loops is reported in Alg. \ref{alg:lazy_const}. 
Intuitively, the algorithm checks the number of tasks covered in the valid path from $0$ to $m+1$. It then compares this number with the total number of tasks covered in the whole schedule. Dissimilarity in these two numbers indicates the existence of an invalid path in the schedule.

\begin{algorithm}
\begin{algorithmic}
\For{\textbf{each} robot $i$}
\State \textit{next} $\gets 0$
\State \textit{count} $\gets 0$
\While{\textit{next} \textbf{is not} $m+1$}
\State \textit{next} $\gets \underset{k}{\arg\max} (x_{i,\textit{next},k}$)
\State \textit{count} $\gets \text{\textit{count}}+1$
\EndWhile
\State \textit{visited} $\gets \sum^{m+1}_{j=0}\sum^{m+1}_{k=0}x_{ijk}$
\If{\textit{count} $\neq$ \textit{visited}}
\State \textbf{return} solution is not valid
\EndIf
\EndFor
\State \textbf{return} solution is valid
\end{algorithmic}
\caption{Detecting loops in a candidate solution}\label{alg:lazy_const}
\end{algorithm}

\subsubsection{Skill allocation}

To satisfy the skills required by the tasks, a robot $i$ must possess at least one of the required skills to attend a task $k \in [1,m]$.
\begin{align}
\forall i \forall k \backslash \{0,m+1\} \quad \sum^{m}_{j=0}x_{ijk} \leq \sum^{l}_{s=0}q_{is}r_{ks}. \label{eq:attnd_req}
\end{align}

For a schedule to be valid, each task must have robots with the required skills. To achieve this, we introduce matrix $Z$, where $z_{ks}$ indicates the number of robots that offer skill $s$ required for task $k$ (Eq. \eqref{eq:arr_skill}). We ensure that each element in $Z$ is greater than or equal to the corresponding element in the skill requirement matrix $R$ (Eq. \eqref{eq:skill_req}).
\begin{align}
\forall s \forall k \backslash \{0,m+1\} &\quad z_{ks} = \sum^{n}_{i=0}\sum^{m}_{j=0} x_{ijk}q_{is} \label{eq:arr_skill}\\
\forall s \forall k \backslash \{0,m+1\} &\quad z_{ks} \geq r_{ks}. \label{eq:skill_req}
\end{align}

The above constraints theoretically allow for a task to have more skills than required. In general this is unavoidable, because the robots contributing to a task might have overlapping skills while also contributing unique ones. However, there is a benefit in avoiding schedules where certain tasks are attended by superfluous robots, i.e., robots that have some of the required skills, but none of them is unique within the coalition. The benefit is that rejecting superfluous robots makes the search space much smaller, significantly reducing run-times as we empirically observed in the experiments we ran during early phases of this work.

To identify superfluous robots, we use the binary matrix $Z^b$ (Eq. \eqref{eq:excess_skill}) where $z_{ks}^b$ equals 1 if skill $s$ is excessive for task $k$, and 0 otherwise.
\begin{equation}
\forall s \forall k \backslash \{0,m+1\} \quad z^b_{ks} = 
\begin{cases}
0                                      & \text{if } z_{ks} \leq r_{ks}\\
1                                      & \text{otherwise}.
\end{cases}
\label{eq:excess_skill}
\end{equation}
We then impose that, if a robot $i$ attends a task $k$, the robot must have at least one skill that is not in excess. In Eq. \eqref{eq:avoid_excess} $z_{ks}^b q_{is}$ is 1 when a skill $s$ of robot $i$ is in excess for task $k$, and $\sum_s z_{ks}^b q_{is}$ is the number of redundant skills robot $i$ has for task $k$. 
Due to constraint \eqref{eq:attnd_req}, if robot $i$ attends task $k$ then $\sum_s q_{is} r_{ks} \geq 1$, i.e., at least one of robot $i$'s skills is required by task $k$. We can then impose the following constraint:

\begin{multline}
\forall i \forall k \backslash \{0,m+1\}  \sum^{m}_{j=0}x_{ijk}=1 \implies \\ \sum^{l}_{s=0} z^b_{ks}q_{is} \leq \bigg(\sum^{l}_{s=0}q_{is}r_{ks}\bigg)-1
\label{eq:avoid_excess}
\end{multline}

\subsubsection{Arrival times}
\label{subsec:arrival_times}
One of the core requirements in a coalition is the simultaneous presence of all its members. In this paper, we assume that the absence of even a single robot makes it impossible for a task to progress. Hence, a task can start when the last required robot has joined the coalition at the location.

To express these requirements, we consider the arrival times of each robot. We introduce matrix $Y$ whose elements $y_{ik}$ store the arrival time of robot $i$ at task $k$. If a robot does not visit a task, its corresponding arrival time is set to 0:
\begin{equation}
\forall i, k \backslash \{0\} \quad  \sum^{m}_{j=0} x_{ijk} = 0 \implies y_{ik} = 0  \label{eq:y_zero}\\
\end{equation}

Task $j$ starts at the arrival time of the last robot to join the coalition, denoted by $y_j^{\text{max}}$.
\begin{equation}
\forall j \backslash \{0\} \quad y^{\max}_j = \underset{i}{\max} (y_{ij})  \label{eq:y_max}
\end{equation}

\begin{figure}[t]
    \centering
    \includegraphics[width=0.45\textwidth]{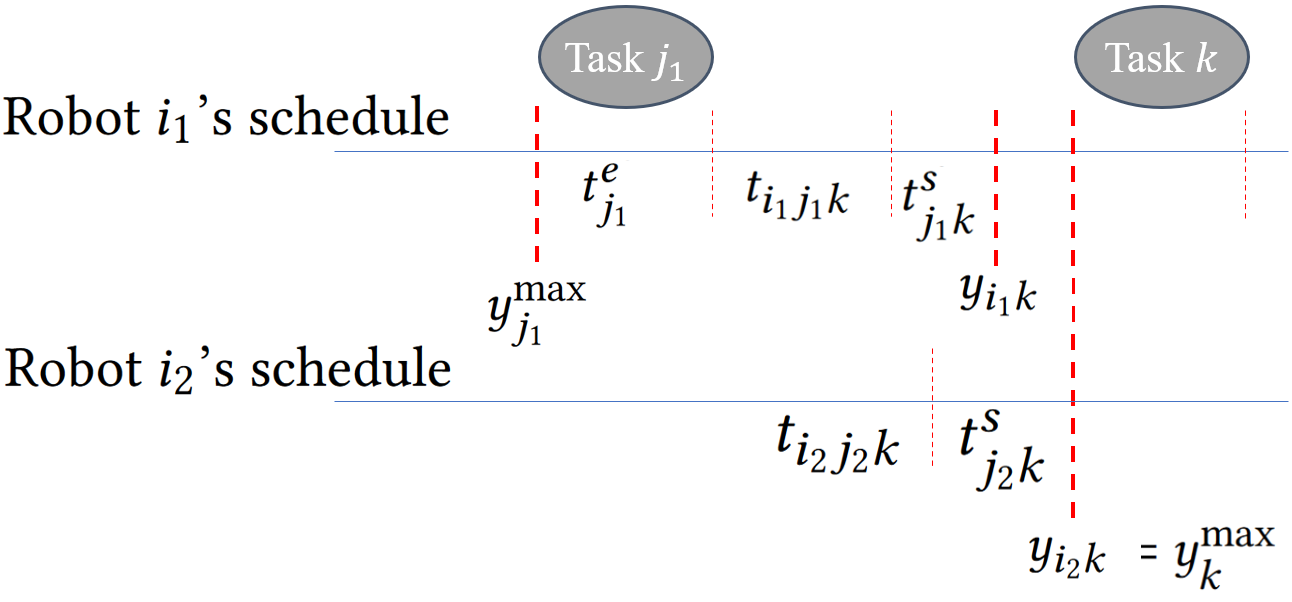}
    \caption{Arrival times of robots $i_1$ and $i_2$. \textmd{Robot $i_1$ attends task $k$ after $j_1$, whereas the robot $i_2$ attends the same task after $j_2$. The duration of execution, travel, and stochastic buffer time are denoted by $t^e$, $t$, and $t^s$, respectively. Task $k$'s starting time is given by $y^{\max}_k$. Robot $i_2$ is the last to arrive at the task, so $y^{\max}_k$ is equal to its time of arrival $(y_{i_1k}$.)}}
    \label{fig:timing_explanation}
\end{figure}

To calculate the arrival time $y_{ik}$ of robot $i$ at a task $k$, it is sufficient to sum the time of completion of the previous task $j$ with the travel time from task $j$ to task $k$ (denoted by $t_{ijk}$). The constraint is then
\begin{multline}
\forall i \forall j \backslash \{m+1\} \forall k \backslash \{0\} \ x_{ijk} = 1 \implies \\ y_{ik} = y^{\max}_{j} + t^{e}_{j} + t_{ijk} + t^s_{jk}
\label{eq:y_value}
\end{multline}
where $t^{e}_{j}$ is the execution time of task $j$ and $y^{\max}_{j}$ indicates the starting time of the same task. The stochastic buffer time between the task $j$ and $k$ is given by $t^s_{jk}$ which will be covered in Sec. \ref{subsec:stoch_time}. A pictorial representation of this calculation is reported in Fig. \ref{fig:timing_explanation}.


\subsubsection{Stochastic travel times}
\label{subsec:stoch_time}
In the quest for a problem formulation that incorporates as many realistic aspects as possible, we include the possibility for travel times to be known only probabilistically. To model travel times as stochastic processes, we assume that the delay can be captured as Gaussian noise $\mathcal{G}(\mu,\sigma)$. More specifically, if we denote with $t$ the ideal travel time between two tasks, then
\begin{equation}
t^* = t + \mathcal{G}(\mu,\sigma).
\end{equation}

For each robot in a coalition, we can express the need to arrive at the task as
\begin{equation}
P(t < \bar{t}) \geq \epsilon \label{eq:prob_constraint}
\end{equation}
where $\bar{t}$ is the hypothetical starting time of the task. We can develop Eq. \eqref{eq:prob_constraint} as follows:
\begin{align*}
t^* = t + \mathcal{G}(\mu,\sigma) &\leq \bar{t} \\
\mathcal{G}(\mu,\sigma) &\leq \bar{t} - t \\
\sigma^2\mathcal{G}(0,1)+\mu &\leq \bar{t} - t\\
P \left( \sigma^2\mathcal{G}(0,1)+\mu \leq \bar{t} - t \right) &\geq \epsilon\\
P \left( \mathcal{G}(0,1) \leq \frac{\bar{t} - t -\mu}{\sigma^2} \right) &\geq \epsilon\\
\Phi \left(\frac{\bar{t} - t -\mu}{\sigma^2} \right) &\geq \epsilon\\
\end{align*}
where $\Phi(\cdot)$ denotes the cumulative distribution function of $\mathcal{G}(0,1)$. Therefore, indicating with $\Phi^{\text{inv}}(\cdot)$ the inverse of $\Phi(\cdot)$, we can write
\begin{align*}
\left(\frac{\bar{t} - t -\mu}{\sigma^2} \right) &\geq \Phi^{\text{inv}}(\epsilon)\\
\bar{t} - t \geq \mu + \sigma^2 \Phi^{\text{inv}}(\epsilon)\\
\end{align*}
This calculation allows us to introduce the symbol $t^s$ defined as follows:
\begin{equation}
t^s = 
\mu + \sigma^2 \Phi^{\text{inv}}(\epsilon)
\label{eq:stoch_final}
\end{equation}
which indicates a ``safety margin'' to arrive on time at a task with probability $\epsilon$ given the mean and standard deviation $\mu, \sigma$ of the road to that task.



\subsubsection{Objective}

The cost function we aim to minimize is the total time taken by the robots to complete the tasks. This corresponds to the arrival time of the last robot at task $(m+1)$. The objective is therefore
\begin{equation}
\text{min} \; y_{m+1}^{\text{max}}
\label{eq:objective}
\end{equation}

\subsubsection{Solving} We use Gurobi \cite{gurobi} to solve the optimization problem. This software is well-known to efficiently produce optimal solutions for convex problems. However, our problem is non-convex, and the objective function (Eq. \eqref{eq:objective}) hints that multiple equally good solutions will exist.

\subsection{Greedy formulation}

The previously discussed method produces an optimal result, but experimental evaluation reveals that it takes a long time to reach a solution. This motivates the need for another method that can solve the same problem quickly, although at the cost of optimality. We propose a simple, but effective greedy solver that produces a quick but sub-optimal result. 


\begin{algorithm}
\begin{algorithmic}[1]
\small

\While{Any task is unsatisfied}

\State $(i_1, k_1),... \gets \text{Robot-task pairs with max contribution}$
\State $(i^c, k^c) \gets \text{The earliest robot-task pair from} (i_1, k_1), ...$

\State  Assign task $k^c$ to the robot $i^c$ using Algorithm \ref{alg:assign_task}

\While{Unaddressed skill at task $k^c$}
\State $i^d_1, ... \gets $Robots with max contributions from remaining skills at $k^c$
\State $i^d \gets $ The earliest robot from $i^d_1, ...$

\State Assign task $k^c$ to the robot $i^d$ using Algorithm \ref{alg:assign_task}
\EndWhile

\State $y^{\max}_{k^c} \gets \underset{i}{\max} \ y_{ik^c}$

\EndWhile
\end{algorithmic}
\caption{The proposed greedy algorithm}\label{alg:greedy_algo}
\end{algorithm}

\begin{algorithm}
\begin{algorithmic}[1]
\small

\State $j \gets$ The current task of robot $i$
\State $x_{ijk}=1$
\State $y_{i,k} \gets y^{\max}_j + t^{e}_{j} + t_{ijk} + t^s_{jk}$
\State Update the list of unaddressed skills at task $k$

\caption{Assign task $k$ to robot $i$}\label{alg:assign_task}
\end{algorithmic}
\end{algorithm}

\subsubsection{Methodology}

In this work, we assume that task execution can only start when all the required skills are fulfilled simultaneously. Thus, a coalition might cause its robots to wait idly until the last robot in the coalition arrives. It is thus desirable to have as small coalitions as possible with robots that cover as many skills as possible. On the other hand, only seeking a solution with small coalitions might require few, powerful robots to spend significant time travelling across the environment to attend the assigned tasks. In such a scenario, the generated paths for the robots are not optimal due to the absence of any mechanism to shorten the robots' travel path. Motivated by these observations, we propose a greedy algorithm that promotes forming small coalitions while also minimizing the distance traveled by the robots.

Our algorithm first finds the robots that can contribute the most to a task and arrive the soonest. We define a robot's `contribution' as the number of previously unoffered skills it can bring to a task. We identify all the robot-task pairs that maximize the robots' contributions to the tasks (Alg. \ref{alg:greedy_algo}, line 2). If multiple robots contribute equally, we choose the one that can reach the task location first (Alg. \ref{alg:greedy_algo}, line 3). This estimated time of arrival is calculated with the same logic as in Sec. \ref{subsec:arrival_times}. 

We use Alg. \ref{alg:assign_task} to add the task to the robot's schedule (line 2), update its arrival time (line 3), and update the task's requirements (line 4) to account for the skills provided by the attending robot. 

We now choose a robot coalition to fulfill the skills required for task $k^c$. If the task still requires additional skills to start (Alg. \ref{alg:greedy_algo}, line 5) we select the robots that can offer the highest number of the remaining skills (Alg. \ref{alg:greedy_algo}, line 6). If multiple robots are tied, we choose the one that can reach the task location first  (Alg. \ref{alg:greedy_algo}, line 7). We then use Algorithm \ref{alg:assign_task} to add the task to the robot's schedule  (Alg. \ref{alg:greedy_algo}, line 8). We repeat this process until all of the task requirements have been fulfilled (Alg. \ref{alg:greedy_algo}, line 5). We then update the task start time for the chosen task $k^c$ according to the attending coalition (Alg. \ref{alg:greedy_algo}, line 10).

\subsubsection{Correctness of the algorithm}
We assert that our algorithm yields a feasible solution in which all tasks are allocated to suitable robots. To establish this claim, we demonstrate that the algorithm assigns each task to a set of appropriate robots. Suppose there is an unassigned task $k$ with unfulfilled requirements, which means the sum of its requirements is greater than 0. The solver must continue until this task is assigned a group of robots that can fulfil all of its requirements. Hence, eventually a robot will choose this task, even if it can only provide a single skill. Once the solver has found a robot for task $k$, it will search for other robots to fulfil any remaining requirements. A feasible solution requires at least one robot to contribute at least one skill to the remaining requirements. As long as such a robot exists, it will be assigned to task $k$. Moreover, the solver will not choose a robot that cannot contribute to the task as long as there is a robot that can contribute at least one skill. Therefore, there can be no redundant robot assigned to any task. We conclude that our algorithm always terminates with a feasible solution. Therefore, we can conclude that the proposed greedy algorithm for task allocation is correct and produces a feasible solution.
\section{Experimentatal Evaluation}

We conducted experiments with 4 robots and 8 tasks, testing 3 configurations with 2, 4, and 8 skills. Each configuration comprised of 30 unique setups that differed in the location of the tasks, their skill requirements, and the allocation of skills among the robots.

\subsection{Experimental Setup}
In our experiments, we define an effective area of $200 \times 200$ units, and we assume that each robot can travel 1 unit of distance per time unit. For each configuration setup, we randomly assign the locations and skill requirements for each task. The task execution time of each task is set uniformly at random from the range $[0,100]$. We also allocate skills to the robots uniformly at random. We set the starting locations of the robots to be evenly distributed around the center of the experiment area. Specifically, the starting location $(p^i_x, p^i_y)$ of the $i^{th}$ robot is calculated using the value $r = 15$ units as follows:

\begin{equation}
    (p^i_x, p^i_y) = \Big(r\  \text{sin} \bigg( \frac{i \pi}{n}\bigg), r\  \text{cos}\bigg(\frac{i \pi}{n}\bigg) \Big).
\label{eq:robot_st_loc}
\end{equation}

We verify the validity of each generated experiment setup by checking the following conditions:

\begin{enumerate}
    \item Each robot is not allocated more than $l/2$ skills; 
    \item Every skill is present at least once in the robot pool;
    \item Every robot possesses at least one skill.
\end{enumerate}

\paragraph{Stochasticity parameters} The value of the mean travel time, denoted by $\mu$, was set to 10\% of the time it takes for the robot to travel between tasks. The value of the standard deviation, denoted by $\sigma$, was set as a random fraction of $\mu$. Specifically, a value was chosen uniformly at random from the interval $[0.05, 0.50]$ and multiplied by $\mu$ to obtain the final value of $\sigma$. This ensured that the amount of variability in the travel times was proportional to the mean travel time. To make the experiment results repeatable, the standard deviation value for each path is assigned at the time of setup generation. This ensures that the same standard deviation values are used throughout all the experiment runs. Finally, the probability of a robot arriving at a task within a given time window, denoted as $\epsilon$, was set to 0.95 to allow for some flexibility in task scheduling.

\paragraph{Computer specifications} The experiments were run on a computing cluster with the following configuration allocation: AMD EPYC 7543 processors, 22 CPU cores, 156 GB of RAM.

\subsection{Discussion}

\begin{figure}[t]
    \centering
    \includegraphics[width=0.45\textwidth]{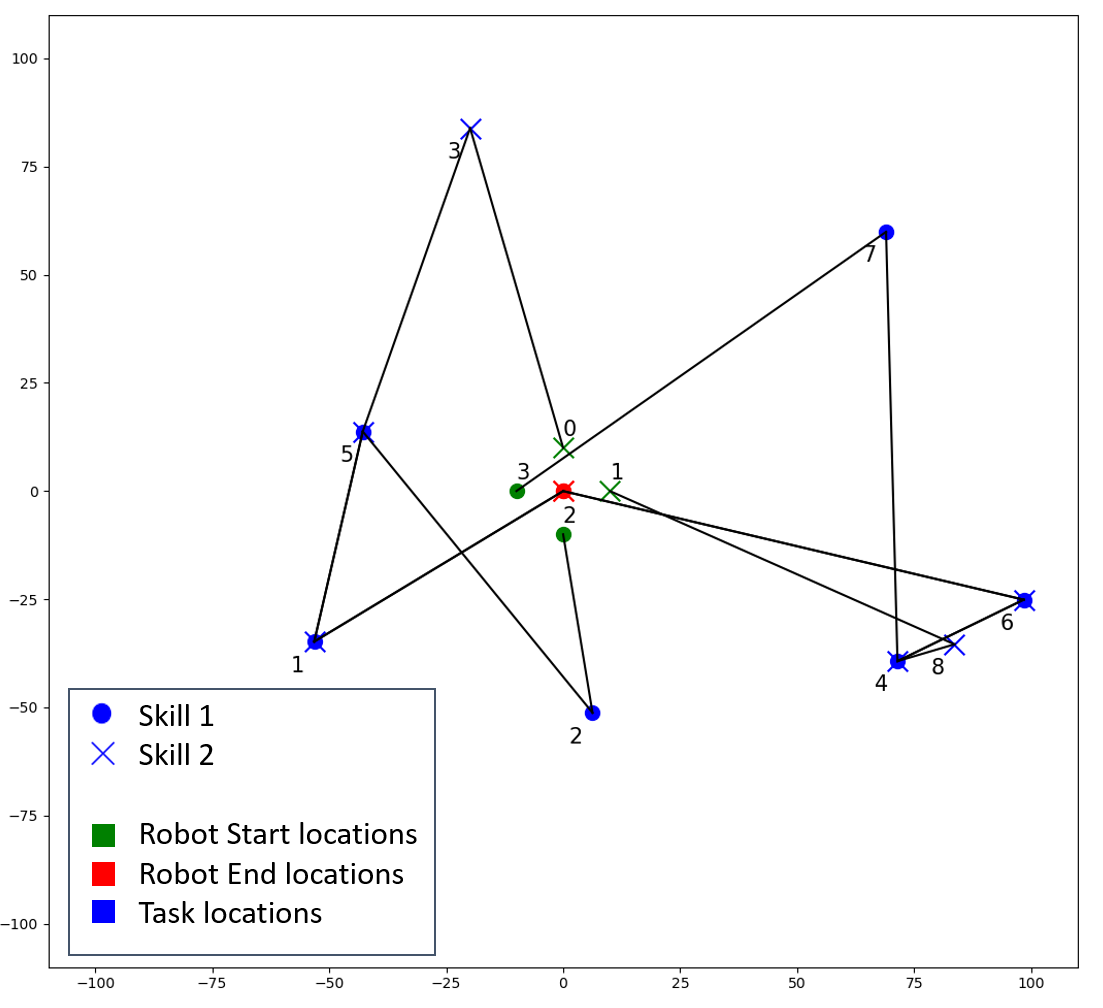}
    \caption{An example solution for a setup with 2 skills, 4 robots, and 8 tasks. The two types of skills are represented by a \textit{cross} and a \textit{circle}. The robots' starting locations are shown by green icons near the center of the area, and their end locations are shown by red icons at the center. If a robot with a \textit{cross} skill starts at a location, a green cross is shown at that location. Blue skill symbols indicate the skills needed for each task at that location, and the tasks are randomly placed in the simulation area. The solid lines show the schedule for each robot, from its starting location to its end location.}
    \label{fig:example_solution}
\end{figure}

To evaluate the performance of the two methods, we analyze two key aspects of the solution. The first one is the final cost of the solution produced. This tells us the quality of the produced solution. The second aspect of interest pertains to the wall clock time required to solve the problem. This allows us to assess the efficiency of the two algorithms in terms of the computational resources and time complexity. A sample of optimal solution for a 2-skill, 4-robot, 8-task setup is reported in Figure \ref{fig:example_solution}.

\begin{figure}
     \centering
     \begin{subfigure}[b]{0.22\textwidth}
         \centering
         \includegraphics[width=\textwidth]{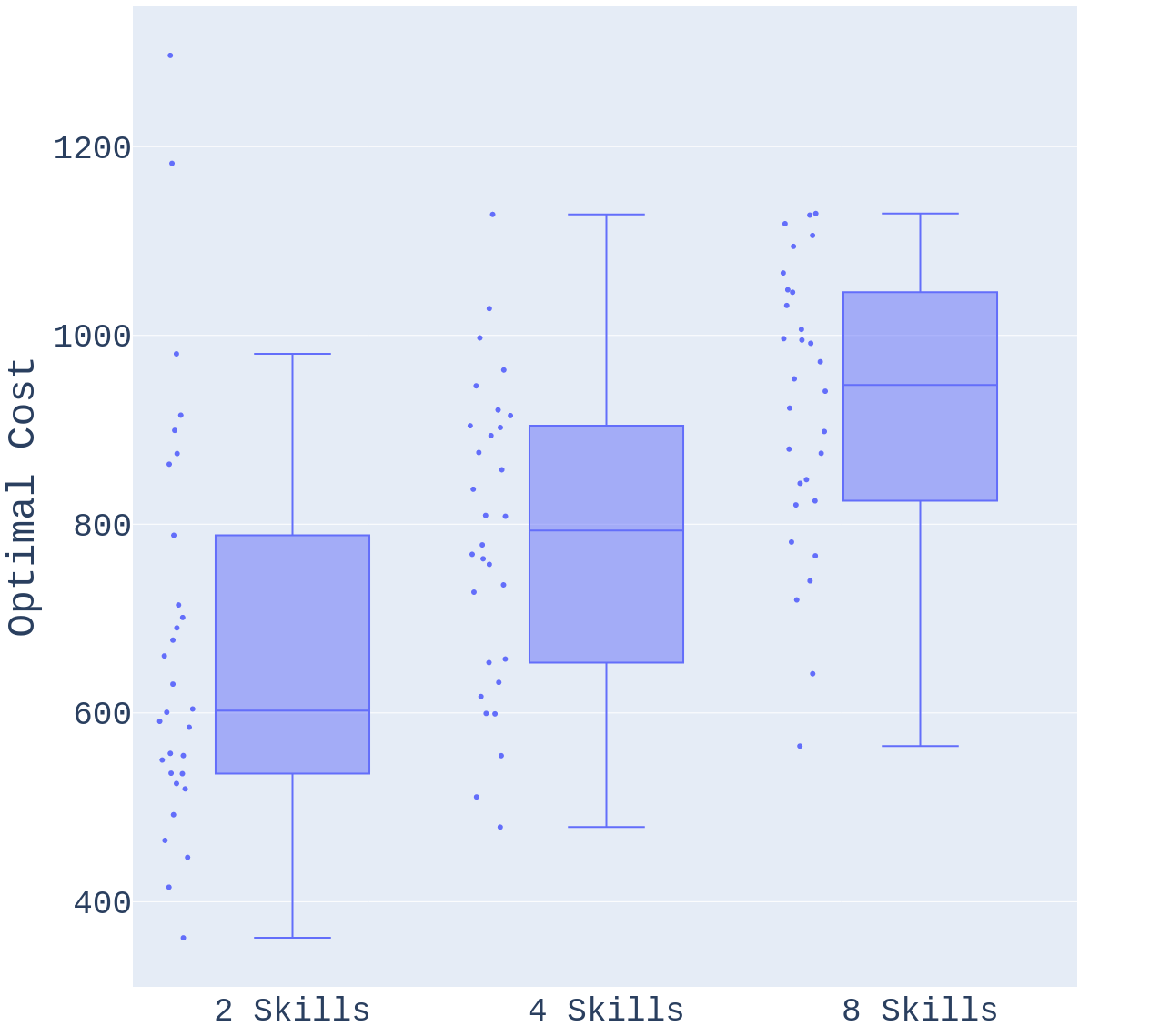}
         \caption{Optimal costs for each configurations}
         \label{fig:optimal_results:cost}
     \end{subfigure}
     \hfill
     \begin{subfigure}[b]{0.23\textwidth}
         \centering
         \includegraphics[width=0.96\textwidth]{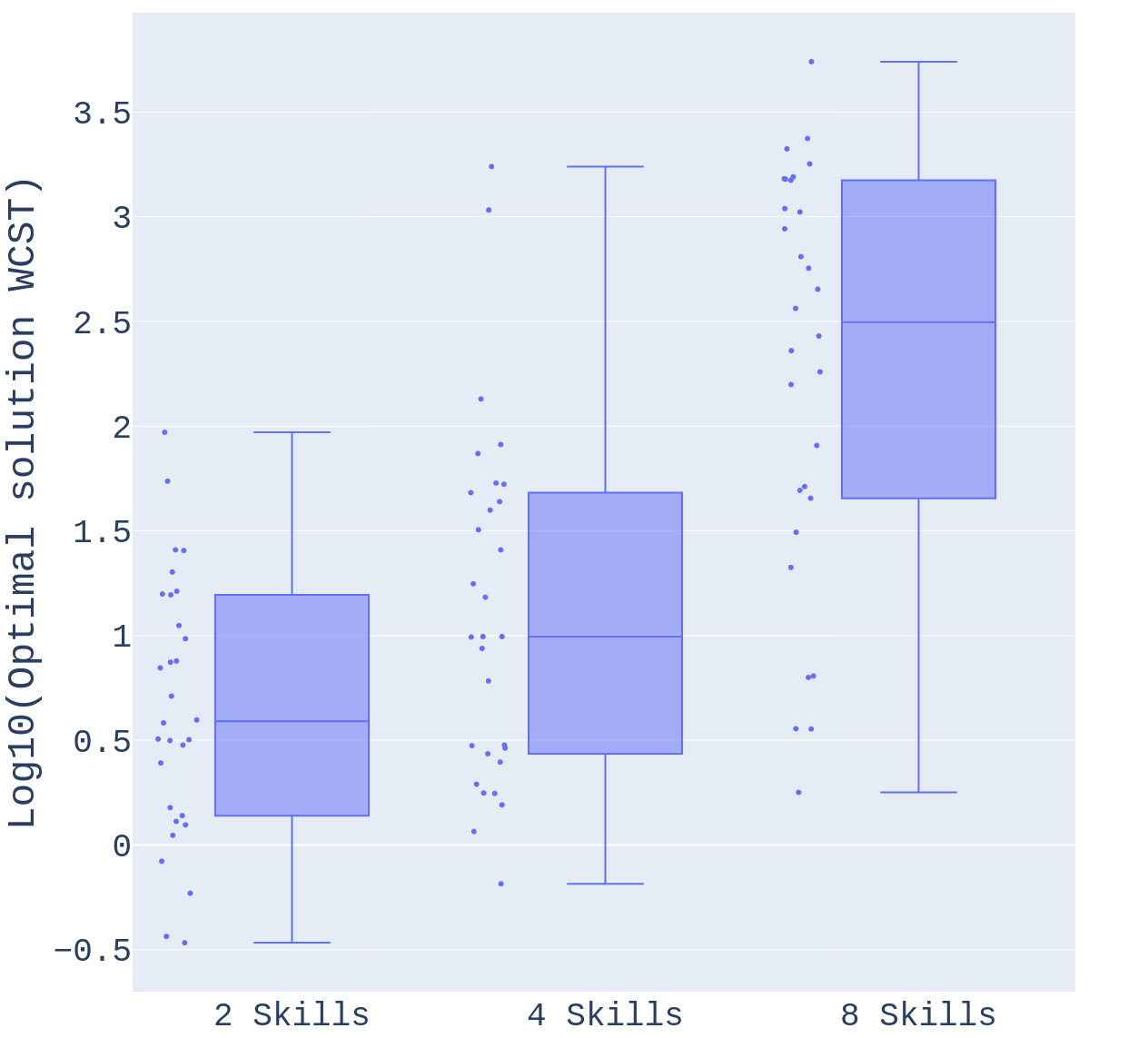}
         \caption{Wall clock solving time (WCST) for each configurations}
         \label{fig:optimal_results:times}
     \end{subfigure}
        \caption{Optimal solution cost and solving times for three different set of configurations with varying number of skills. We can see that as the number of skills increase, both the solution cost and the solving time increases.}
        \label{fig:optimal_results}
\end{figure}

\subsubsection{Optimal solver}
Figure \ref{fig:optimal_results} presents the results for the optimal solver. As we double the total number of skills required, both the solution cost and the wall clock solving time (WCST) increase. However, the notable increase is in the WCST of the 8-skilled setups. This indicates a significant increase in the computational resources required to solve the problem at higher scale. In this configuration, some setups required about 2,000 seconds and one of the setup required 5,000 seconds to declare the final solution as optimal. Based on these results, it would not be realistic to solve a problem larger than the setup presented, as the computational resources required would be prohibitively high.

\subsubsection{Greedy solver}

\begin{figure}[t]
     \centering
     \begin{subfigure}[b]{0.22\textwidth}
         \centering
         \includegraphics[width=\textwidth]{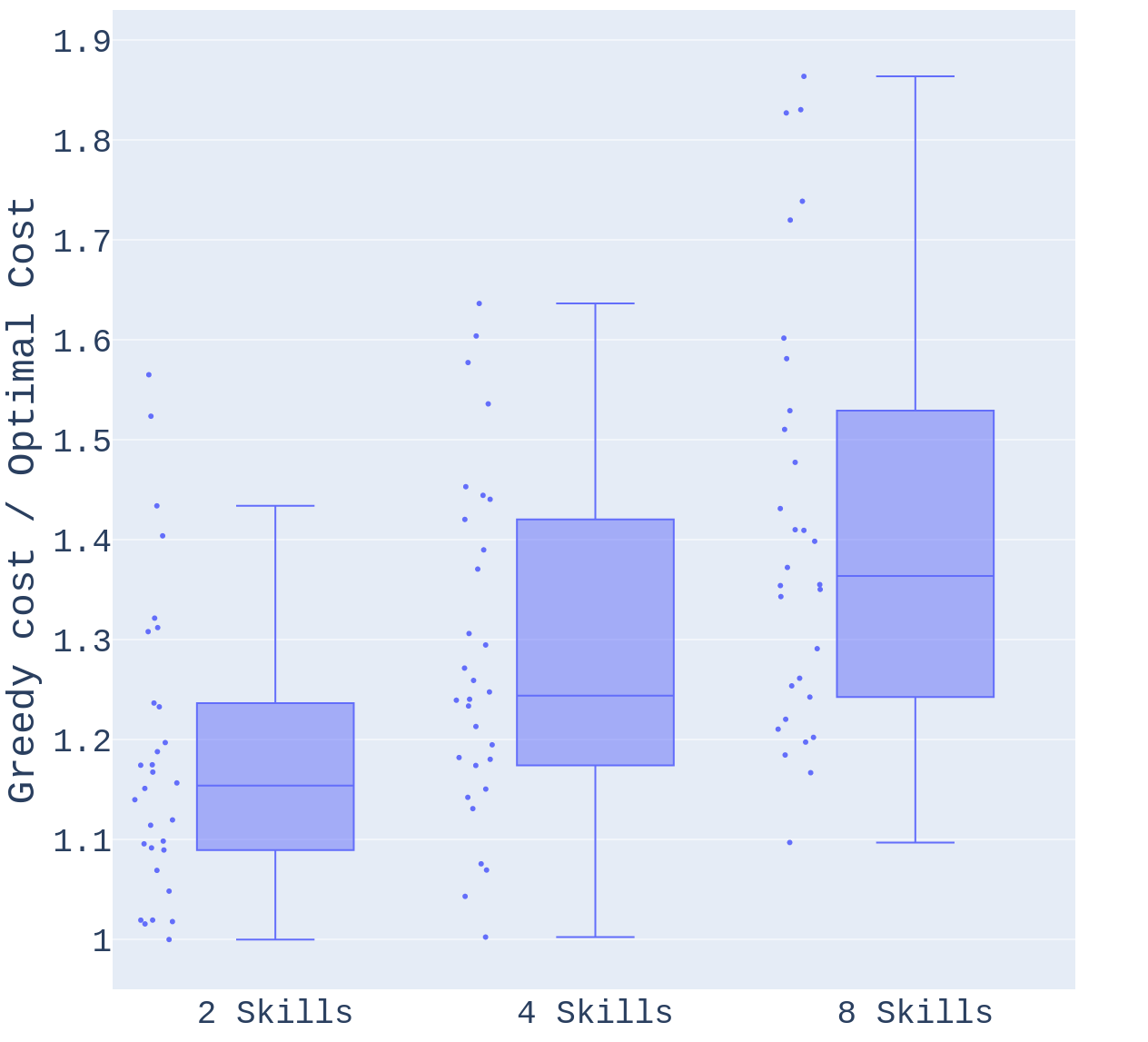}
         \caption{Solution quality of the greedy solver}
         \label{fig:greedy_results:cost}
     \end{subfigure}
     \hfill
     \begin{subfigure}[b]{0.22\textwidth}
         \centering
         \includegraphics[width=\textwidth]{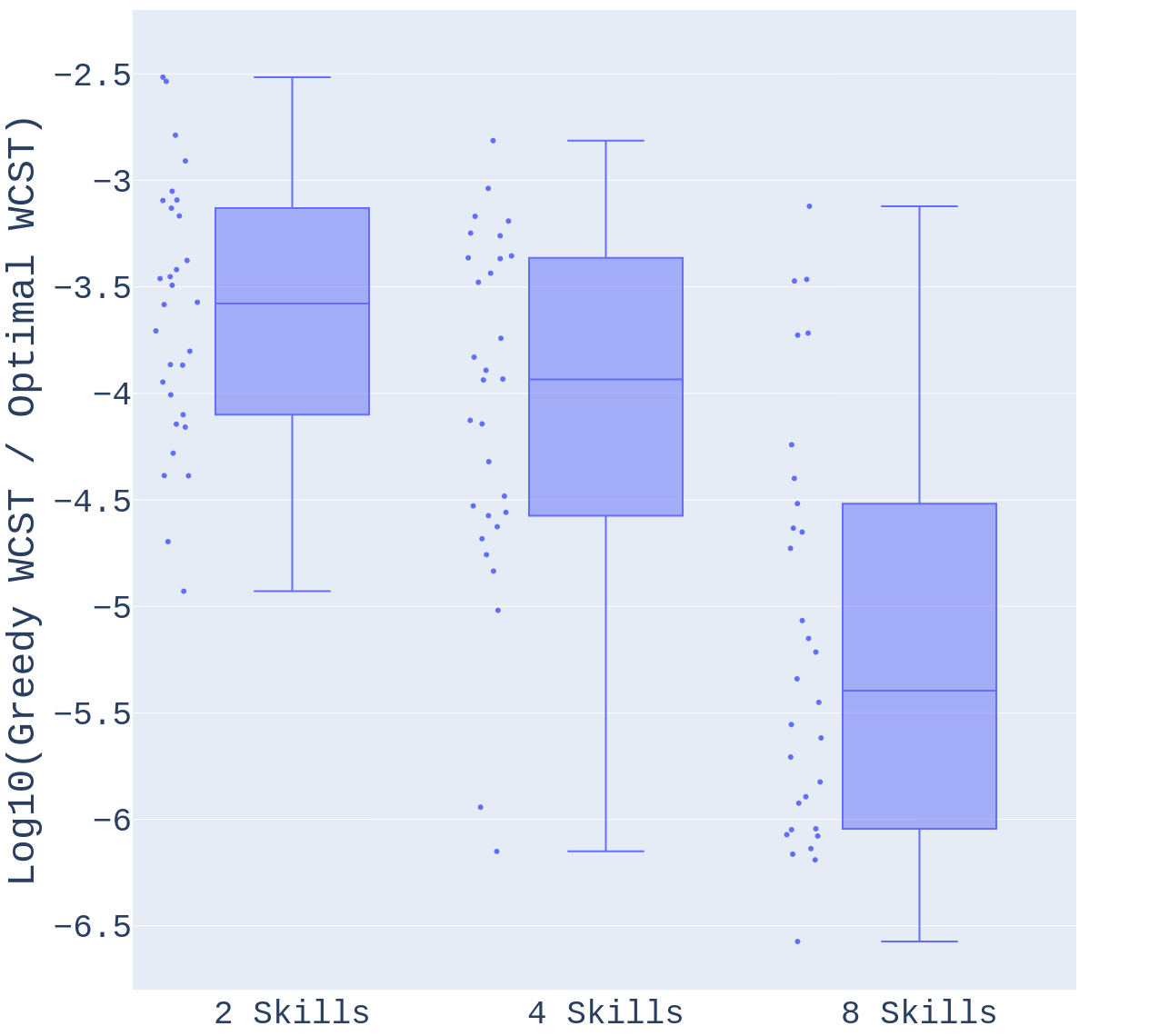}
         \caption{Relative speed of the greedy solver}
         \label{fig:greedy_results:times}
     \end{subfigure}
        \caption{Performance of the greedy solver as compared with the optimal solver. We can see that the greedy solution gets further away from the optimum solution as we increase the number of skills. But it takes the greedy solver orders of magnitude lesser amount of time, as compared with the optimum solver}
        \label{fig:greedy_results}
\end{figure}
To analyze the greedy solver, we compared its performance with the optimal solver. Figure \ref{fig:greedy_results} compares the performance of the greedy solver to that of the optimal one. In Figure \ref{fig:greedy_results:cost} we can see that for the configuration with 2 skills, multiple points lie near 1. This means that most of the solutions were very close to the optimum. The median relative cost performance of the greedy solver was 1.15.

The greedy solver performs very well in terms of computational efficiency, as shown in Figure \ref{fig:greedy_results:times}. In 2-skills setups, the median $\log_{10}$ relative run-time is -3.57, indicating that the greedy solver is more than three orders of magnitude faster than the optimal solver. These results showcase the fast nature of the greedy solver and its potential for use in scenarios where real-time decision-making is required. 


As we scale up the problem, the performance of the greedy solver in terms of cost slightly degrades. For the configuration with 8 skills, none of the experiments produced a solution close to the optimum. As indicated by the median cost performance of 1.36, the greedy solutions are off the optimal solution by a significant margin. However, as shown in Figure \ref{fig:greedy_results:times}, most solutions generated using the greedy solver are produced within a factor of $10^{-5}$ of the time it took to solve the same problem using the optimal method, demonstrating the efficiency of the greedy solver. Although the quality of the solutions generated by the greedy solver may not be very good for larger problem sizes, the method is extremely fast and can be useful for scenarios where quick, ``good enough'' decision-making is prioritized over solution optimality.

\begin{figure}[t]
     \centering
     \begin{subfigure}[b]{0.22\textwidth}
         \centering
         \includegraphics[width=\textwidth]{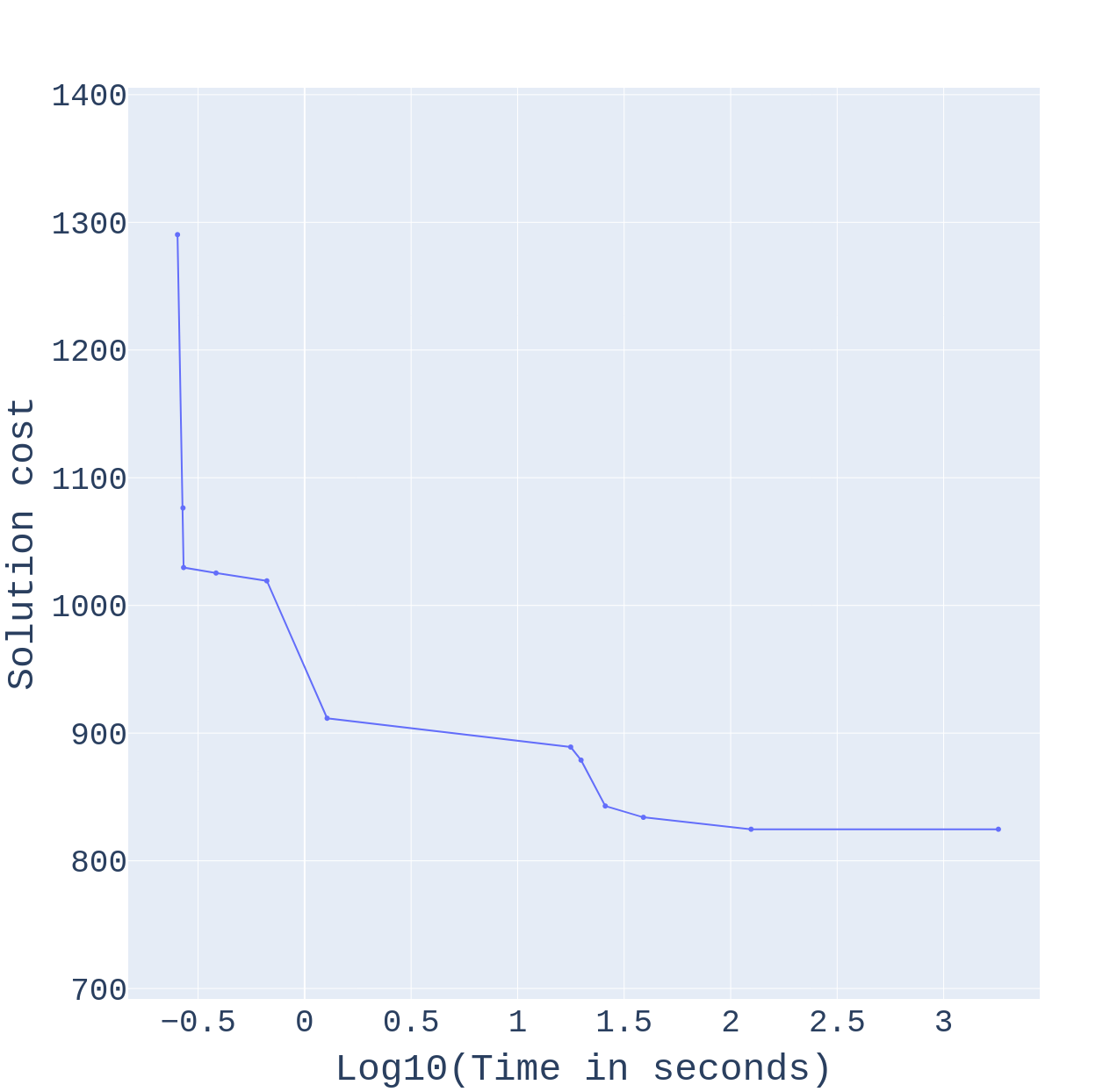}
         \caption{Solution cost progression by the optimum solver for a typical setup of 8 skills}
         \label{fig:gurobi:progression}
     \end{subfigure}
     \hfill
     \begin{subfigure}[b]{0.22\textwidth}
         \centering
         \includegraphics[width=\textwidth]{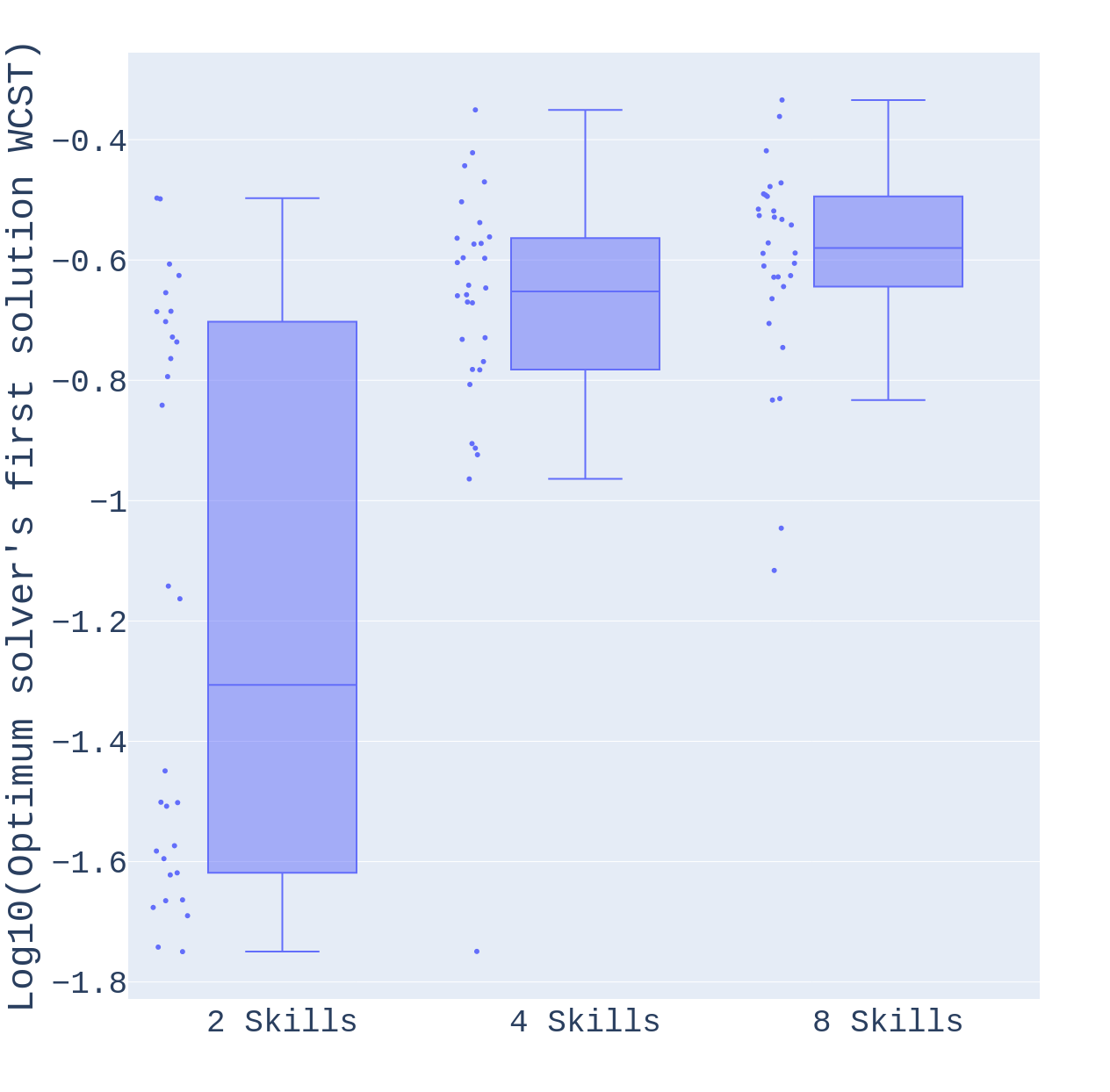}
         \caption{Time taken to produce the first feasible solution by the optimum solver}
         \label{fig:gurobi:first_solution}
     \end{subfigure}
        \caption{Progression of the solution cost produced by the optimum solver. The solver finds a sub-optimal solution quickly, but then spends long time improving and proving the solution as the optimal.}
        \label{fig:gurobi}
\end{figure}

\subsubsection{Optimal solver's first solution}

It is interesting to analyze the time taken by the optimal solver to reach its first solution, neglecting the time needed to verify whether it is optimal. As shown in Figure \ref{fig:gurobi:progression}, for the configuration with 8 skills, the solver rapidly produced a feasible but sub-optimal solution. However, it took a significantly longer time to reach the optimal solution and even more time to prove its optimality. This is further illustrated in Figure \ref{fig:gurobi:first_solution}, which displays the time taken by the solver to obtain the first feasible solution. The optimal solver consistently produces the first solution quickly, but it takes a long time to reach the optimum and prove its optimality (Figure \ref{fig:optimal_results:times}). 

\begin{figure}[t]
     \centering
     \begin{subfigure}[b]{0.22\textwidth}
         \centering
         \includegraphics[width=\textwidth]{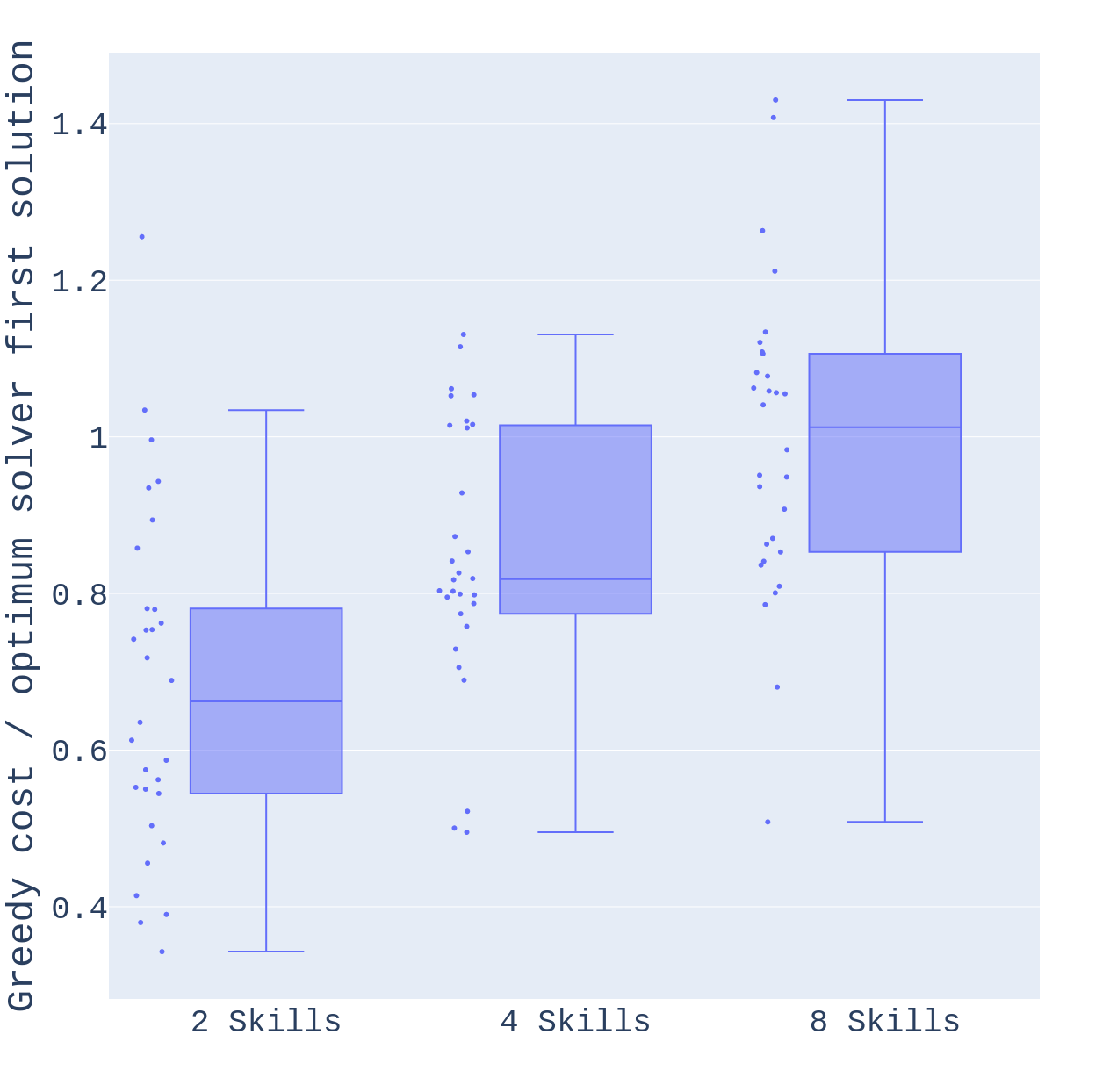}
         \caption{Comparing greedy solver's and optimum solver's first solution costs}
         \label{fig:greedy_fs_compare:cost}
     \end{subfigure}
     \hfill
     \begin{subfigure}[b]{0.22\textwidth}
         \centering
         \includegraphics[width=\textwidth]{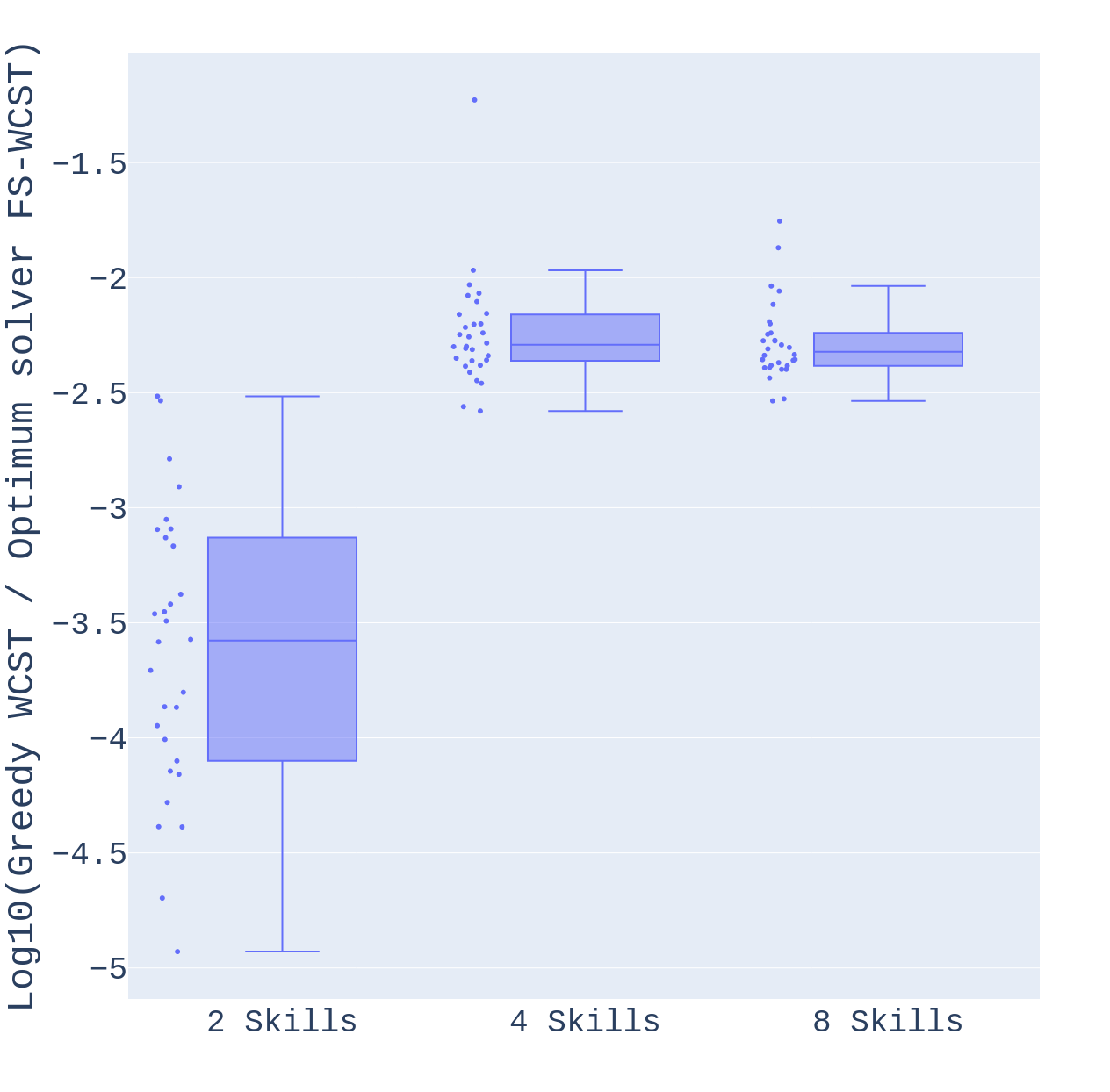}
         \caption{Comparing greedy solver's and the optimum solver's first solution (FS) WCST}
         \label{fig:greedy_fs_compare:times}
     \end{subfigure}
        \caption{Performance of the greedy solver as compared with the optimal solver's first solution. We can see that the greedy solutions are mostly better than the first solutions offered by the optimum solver. The greedy solver is also orders of magnitude faster, as compared with the optimum solver.}
        \label{fig:greedy_fs_compare}
\end{figure}

We compared the performance of the greedy solver with the first solution offered by the optimal solver in Figure \ref{fig:greedy_fs_compare}. As shown in Figure \ref{fig:greedy_fs_compare:cost}, most of the solutions produced by the greedy solver are better than the first solutions provided by the optimal solver. While the optimal solver's solutions are superior half of the times with the 8-skill configurations, the greedy solver is still faster. The greedy solver is consistently more than two orders of magnitude faster than the optimal solver, as shown in Figure \ref{fig:greedy_fs_compare:times}. The data clearly shows the significant speed advantage of the greedy solver over the optimal solver.

\subsubsection{Large-scale experiments}
To further investigate the speed and scalability of the proposed methods, we conducted a series of experiments on larger-scale configurations. We set the number of robots at 32 and the number of skills at 64, and generated 30 setups for each of four task counts: 128, 256, 512, and 1,024. For such large scales, the optimal solver failed to produce even the first solution after running for three hours. Hence, we analyze only the greedy solver's performance in what follows.

\begin{figure}[ht]
    \centering
    \includegraphics[width=0.45\textwidth]{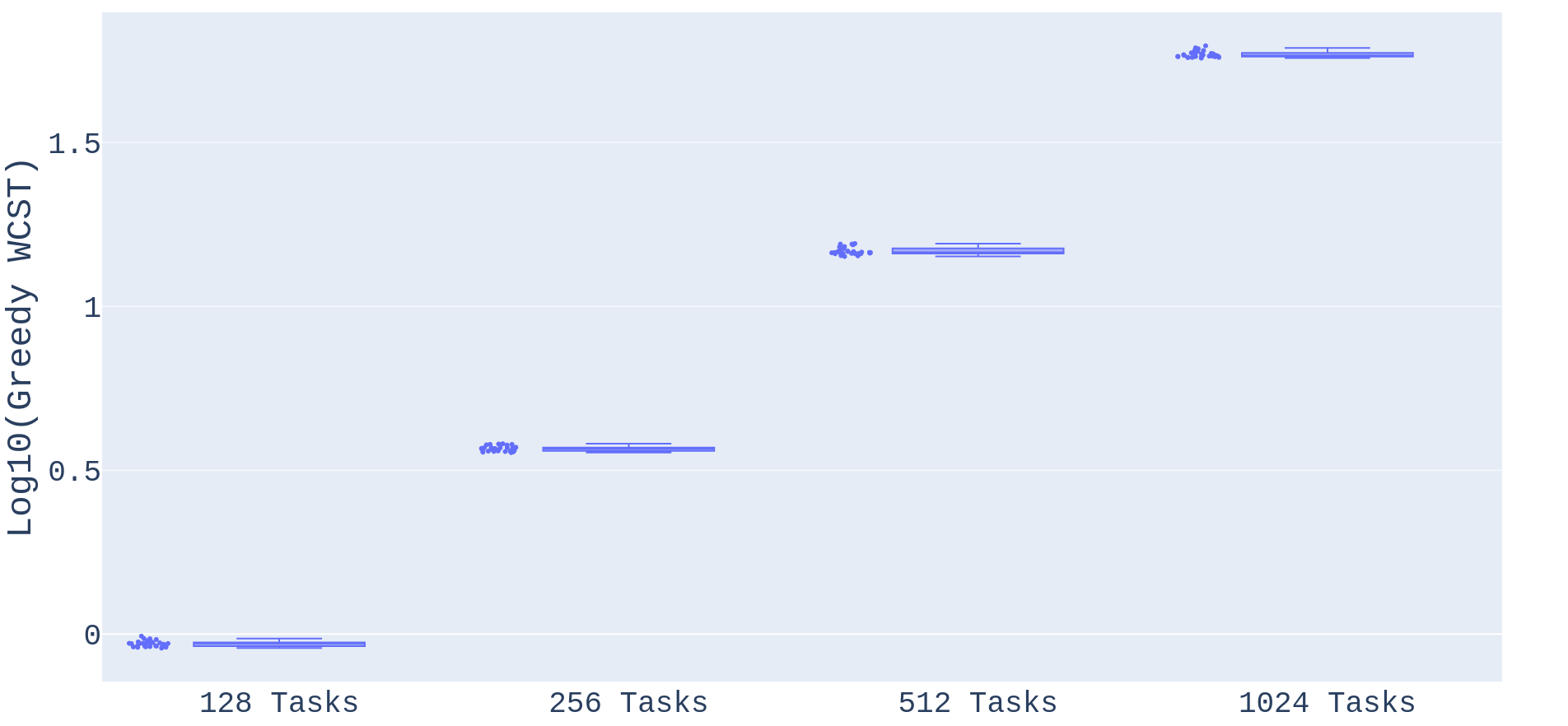}
    \caption{Wall clock solving times (WCST) required by the greedy solver for large set of tasks}
    \label{fig:large_results}
\end{figure}

Figure \ref{fig:large_results} displays the $\log_{10}$ of the wall clock times required by the greedy solver to solve each of these configurations. 
The results reveal that the $\log_{10}$ of the WCST required to solve the larger-scale configurations increases by approximately 0.6 for each doubling of the number of tasks. In other words, as the number of tasks doubles, the solve times increase by a factor of approximately four. Despite this increase, we consistently obtained solutions for the configuration of 1,024 tasks within 45-50 seconds. These results suggest that the greedy solver can handle large-scale instances efficiently, making it a promising approach for real-world scenarios with a large number of tasks.

\section{Conclusion}
In this work, we presented an approach to task allocation in heterogeneous multi-robot systems. Our problem combines coalition and scheduling of a heterogeneous swarm of multi-skilled robots. Our problem formulation also includes stochastic aspects of travel between any two tasks. We proposed two methods to solve this problem. The first produces an optimal solution at the expense of long run-times. This method is onsly suitable for small-scale problems where optimality is required. Our second proposed method uses a greedy approach and it quickly produces sub-optimal solutions.

We compared the performance of the two methods. We found that the greedy solver is typically between 2x the cost of the optimal solution, but it offers speedups in the order of $10^{5}$ with respect to the optimal solver. Further, the greedy solver can tackle large-scale scenarios (32 robots, 64 skills, and 1,024 tasks) in less than a minute. This makes the greedy solver a viable option for quick but best-effort decision-making.

In future work, we aim to improve the performance of the greedy solver by using better heuristics. We also aim to make the system decentralized to promote parallelism.

\bibliographystyle{IEEEtran}
\bibliography{references}
\balance


\end{document}